%% file: neurips_2026.tex
\documentclass{article}

\usepackage[preprint]{neurips_2026}

\usepackage[utf8]{inputenc} 
\usepackage[T1]{fontenc}    
\usepackage{hyperref}       
\usepackage{url}            
\usepackage{booktabs}       
\usepackage{amsfonts}       
\usepackage{nicefrac}       
\usepackage{microtype}      
\usepackage{xcolor}         
\usepackage{adjustbox}
\usepackage{multirow}
\usepackage[table]{xcolor}
\usepackage{wrapfig}
\usepackage{amsmath}
\usepackage{cleveref}  
\usepackage{amssymb}   
\usepackage{caption}
\definecolor{bestblue}{HTML}{C6E2F2}  
\definecolor{secondblue}{HTML}{EBF5FB}

\title{WAT3R: Feedforward Underwater 3D Reconstruction}

\author{
  Jiayi Xu\textsuperscript{1},
  Jiahao Lu\textsuperscript{1},
  Ziqiang Zheng\textsuperscript{1},
  Yihao Tan\textsuperscript{2}, Yaolong Zhu\textsuperscript{3}   \\
  \textbf{
  Yuan Liu\textsuperscript{1},
  Sai-Kit Yeung\textsuperscript{1}}
  \\
  \textsuperscript{1}The Hong Kong University of Science and Technology \\
  \textsuperscript{2}The Chinese University of Hong Kong \\
  \textsuperscript{3}Peking University \\
  \\[0.5pt]
  \url{ https://xujiayi777.github.io/WAT3R.github.io/}
}

\begin{document}

\maketitle
\input{Sec/0_abstract}    
\input{Sec/1_intro}   
\input{Sec/2_rw}
\input{Sec/3_method}

\input{Sec/4_experiment}

\input{Sec/5_conclusion}

\bibliographystyle{unsrt}
\bibliography{main}


\appendix
\input{Sec/X_suppl}


\end{document}

%% file: Sec/0_abstract.tex
\begin{figure*}[!ht]
    \centering
    \vspace{-20pt}
    \includegraphics[width=0.95\textwidth]{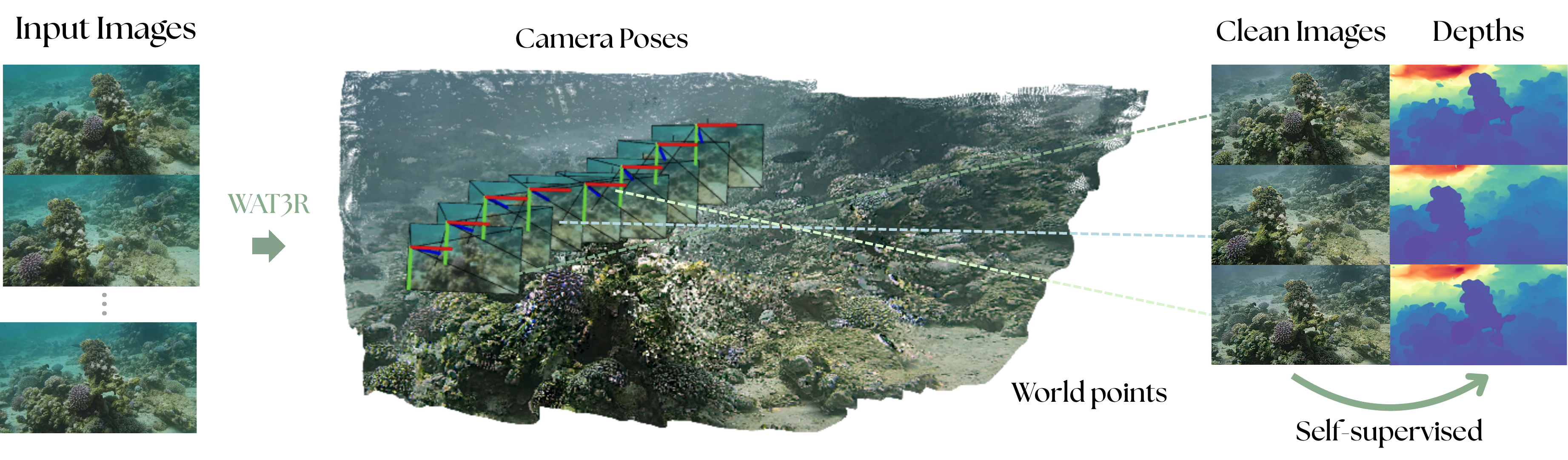}
    \caption{Given degraded underwater images (left), \textbf{WAT3R} performs reliable 3D reconstruction under degradation by directly estimating scene geometry in a single forward pass (middle). It uses self-supervised geometry-driven degradation adaptation to enable consistent feature matching, producing a accurate 3D scene with improved visual quality (right).}
    \label{fig:teaser}
\end{figure*}

\begin{abstract} 
Reliable feedforward underwater 3D reconstruction remains challenging due to severe light attenuation and backscattering, which degrade visual quality and disrupt feature consistency across views, leading to inaccurate multi-view geometry. To address this issue, we propose \textbf{WAT3R}, a feed-forward framework for reconstructing 3D scenes directly from underwater images. By leveraging degradation adaptation as a geometry-constrained process, WAT3R integrates a lightweight neural adaptation module to flexibly account for these underwater imaging effects, thereby improving multi-view reconstruction quality. Implemented in a single forward pass, WAT3R directly and efficiently outputs pixel-aligned 3D point maps and camera poses from underwater videos, allowing a high-quality underwater 3D reconstruction. Experiments conducted on the FLSea, SQUID, and USOD10K datasets show that our method consistently outperforms state-of-the-art approaches on 3D reconstruction tasks, including multi-view/monocular depth estimation and camera pose estimation.
\end{abstract}


%% file: Sec/1_intro.tex
\section{Introduction}

Accurate underwater 3D reconstruction is a fundamental capability for autonomous underwater systems, supporting applications such as navigation~\cite{Rahman2018} and inspection~\cite{Wang2025inspection}. Despite recent advances in 3D reconstruction in terrestrial environments, achieving reliable underwater 3D reconstruction remains challenging due to the unique optical properties of water. In particular, underwater imaging is governed by wavelength-dependent attenuation and backscattering~\cite{Akkaynak2017}, which are inherently geometry-dependent. The magnitude of degradation is determined by optical path length, surface orientation, and viewing direction, making the observed radiance a direct outcome of the interaction between light transport and scene geometry. As a result, the reliability of feature matching and geometric consistency across views is significantly reduced, making accurate geometry estimation a central challenge in underwater 3D reconstruction.

Existing methods for underwater 3D reconstruction~\cite{levy2023seathrunerf, sethuraman2023waternerf,Zhou2024waterhenerf,li2025watersplatting,yang2025seasplat,zhang2026watercleargs} are usually slow and unreliable, while recent feedforward 3D reconstruction approaches~\cite{wang2025vggt, wang2025pi, lin2025DA3, keetha2026mapanything} are promising due to their high efficiency and quality. Most existing underwater pipelines follow a two-stage paradigm: they first estimate camera poses using traditional Structure-from-Motion (SfM), and then perform per-scene optimization for dense 3D reconstruction. SfM requires high-quality image matching, which is challenging for the underwater environment with degraded image quality. Moreover, the per-scene dense 3D optimization is computationally intensive, restricting real-time applicability and scalability. Recent feed-forward ViT-based approaches~\cite{wang2025vggt, wang2025pi, keetha2026mapanything, lin2025DA3} developed for terrestrial scenes enable efficient, scene-agnostic reconstruction without iterative optimization. For example, VGGT~\cite{wang2025vggt} and Pi3~\cite{wang2025pi} directly infer all key 3D attributes of a scene given a set of RGB images. These advances suggest a promising direction: extending feedforward frameworks to underwater scenarios, potentially overcoming the limitations of traditional pipelines. 


Motivated by this, we design \textbf{WAT3R}, a feed-forward framework for underwater 3D reconstruction in this paper. WAT3R allows us to directly estimate both camera poses and dense 3D points in a feedforward manner from an underwater video, which is much more efficient than previous optimization-based methods. Meanwhile, by implicitly learning the 3D prior for underwater scenes, WAT3R overcomes the image quality degradation, resulting in a better 3D reconstruction quality.

However, developing such a feedforward framework for underwater scenes is non-trivial. Directly applying terrestrial pre-trained models to underwater videos inevitably suffers from a severe domain gap. These models fail to generalize to the complex optical distortions caused by wavelength-dependent attenuation and backscattering, leading to significantly degraded reconstruction accuracy. A straightforward solution would be to fine-tune these models; however, this reveals the most significant bottleneck in underwater 3D vision: the extreme scarcity of training data with ground-truth geometry. Acquiring accurate, large-scale 3D ground truth in underwater environments is prohibitively expensive and practically unfeasible. Consequently, it is imperative to explore self-supervised learning strategies that can adapt powerful terrestrial pre-trained models to underwater domains without relying on inaccessible ground-truth annotations.

To overcome these challenges, we propose a physics-guided two-stage adaptation framework for underwater geometry estimation. Our key idea is to leverage the physical insights of the Underwater Image Formation Model (UIFM)~\cite{Akkaynak2018} to bridge the domain gap between terrestrial and underwater imagery. In the first stage, we synthesize pseudo-underwater data by applying UIFM-based optical degradations such as wavelength-dependent attenuation and backscattering to large-scale terrestrial datasets with available ground-truth geometry. This enables the network to learn underwater-specific visual characteristics while still being supervised by accurate geometric labels. In addition, we incorporate a UIFM-based self-supervised reconstruction constraint at this stage: the network jointly predicts depth and a clean image, and is trained to satisfy the physical imaging relationship defined by the UIFM. In the second stage, we further adapt the model to real underwater data using a purely self-supervised learning strategy. To robustly handle occlusions and violations of static-scene assumptions, we adopt a minimum reprojection loss with auto-masking inspired by~\cite{Godardmonodepth2}. Specifically, the model enforces photometric consistency across multiple views in the clean image, while ignoring pixels that do not satisfy geometric consistency. This stage refines the network’s predictions directly on real underwater inputs without requiring ground-truth depth, leading to improved generalization and reconstruction accuracy in complex underwater environments.

We conduct extensive experiments on multiple underwater datasets to evaluate the effectiveness of WAT3R, including multi-view, monocular depth estimation, and camera pose estimation. The results demonstrate that our method consistently outperforms recent feed-forward underwater 3D reconstruction approaches, while maintaining superior geometric fidelity under underwater scenarios. 

%% file: Sec/2_rw.tex
\section{Related Work}
\textbf{Optimization-based Reconstruction.}
Underwater scene reconstruction has progressed toward physics-informed neural rendering~\cite{Akkaynak2019seathru} that unifies optical modeling~\cite{Tang2024} and 3D estimation~\cite{zhang2026watercleargs}, incorporating light transport physics to achieve more robust and consistent results across diverse underwater conditions. NeRF-based methods~\cite{Tang2024, levy2023seathrunerf, sethuraman2023waternerf,Zhou2024waterhenerf} model wavelength-dependent backscattering and attenuation during scene reconstruction. WaterHE-NeRF~\cite{Zhou2024waterhenerf} predicts illuminance attenuation using histogram-equalized pseudo-ground truth, while SeaThru-NeRF~\cite{levy2023seathrunerf} separates direct transmission from backscatter. These approaches are accurate but computationally heavy and scene-specific. 3DGS-based methods are more efficient. WaterSplatting~\cite{li2025watersplatting} augments Gaussian primitives with volumetric water properties, SeaSplat~\cite{yang2025seasplat} predicts separate backscattering and attenuation maps, and WaterClear-GS~\cite{zhang2026watercleargs} adopts a dual-branch design for joint photometric consistency and water-free appearance restoration. However, these methods still require per-scene optimization, limiting scene-agnostic inference and reconstruction accuracy. 

\textbf{Feedforward Reconstruction.} Recently, feed-forward models have emerged as a powerful alternative to optimization-based pipelines, enabling direct prediction of scene geometry from multiple images in a single forward pass. Early methods~\cite{Wang2024dust3r,zhang2024monst3r,lu2025align3r} such as DUSt3R~\cite{Wang2024dust3r} estimate pairwise 3D point clouds in a reference camera frame, but require an additional global alignment stage for large-scale scenes, which can be costly and unstable. Subsequent works address these limitations by improving scalability and simplifying the reconstruction process: Fast3R~\cite{yang2025fast3r} enables joint inference over thousands of images without explicit alignment, while FLARE~\cite{zhang2025flare} decomposes the problem by first estimating camera poses and then reconstructing scene geometry. Building upon these advances, recent feed-forward transformer-based models further push the limits of generalization and efficiency, where ViT-style architectures~\cite{wang2025vggt, wang2025pi, lin2025DA3, keetha2026mapanything} leverage large-scale training and multi-task learning to achieve scene-agnostic 3D reconstruction with real-time inference across diverse in-air scenarios. However, directly applying these models to underwater environments remains challenging due to severe image degradation caused by wavelength-dependent backscattering and  attenuation. To address this, SeaVGGT~\cite{zhao2026seavggt} introduces a self-supervised framework that models depth-dependent attenuation via a graph of learnable prototypes, enabling robust geometry estimation without requiring annotations. Despite this progress, a feed-forward framework that implicitly leverages a degradation adaptation signal to enhance 3D geometry remains underexplored.


%% file: Sec/3_method.tex
\section{Method}
\begin{figure}[t]
  \centering
  \includegraphics[width=\textwidth]{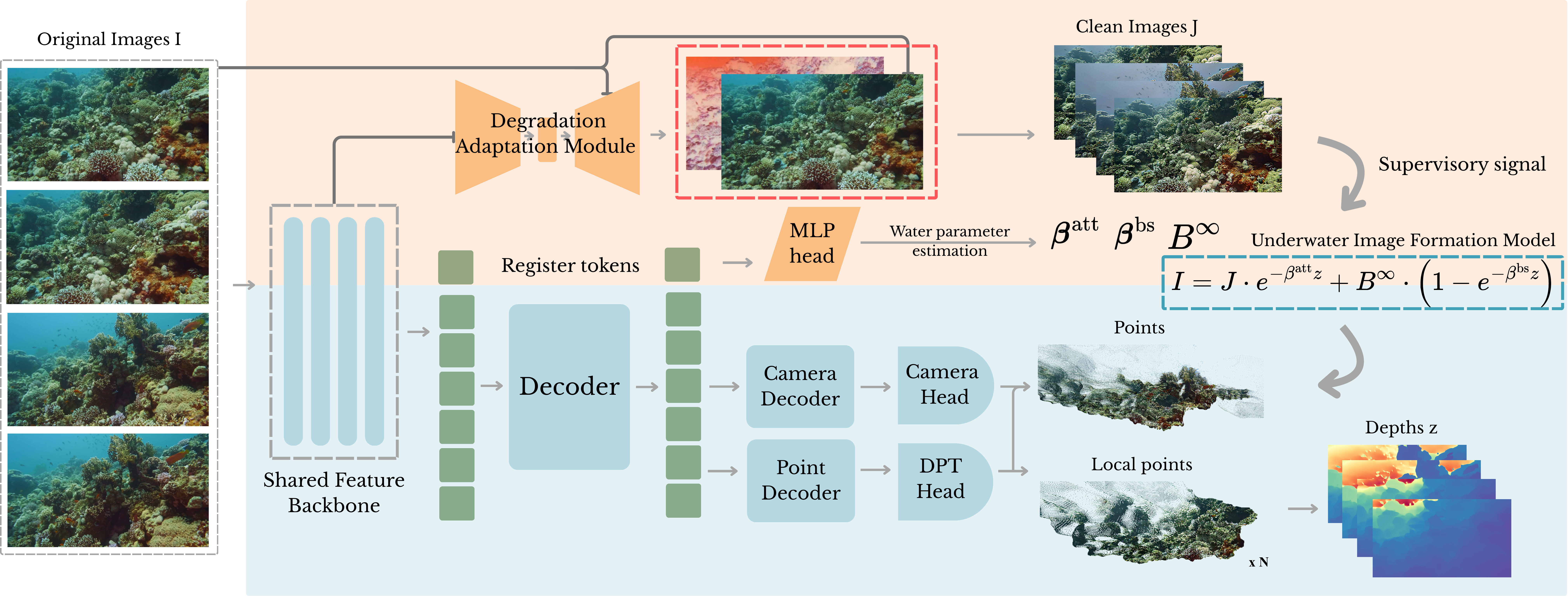}
  \caption{Overview of the WAT3R framework. Our method integrates a degradation adaptation module with a 3D reconstruction pipeline. By incorporating an underwater image formation model, WAT3R utilizes degradation-aware adaptation to implicitly refine depth estimation and 3D point cloud estimation from original images.}
  \label{fig:pipeline}
  \vspace{-10pt}
\end{figure}

\textbf{Overview.}
Given a sequence of $N$ underwater images $\{\mathbf{I}_i \in \mathbb{R}^{H \times W \times 3} \mid i = 1, \ldots, N\}$, our goal is to jointly estimate: (1) the pixel-aligned 3D point maps $\{\mathbf{P}_i \in \mathbb{R}^{H \times W \times 3}\}_{i=1}^N$ in the camera coordinate; (2) and the relative camera poses $\{\mathbf{T}_i \in \mathrm{SE}(3)\}_{i=1}^N$.

\begin{figure}[t]
  \centering
  \includegraphics[width=\textwidth]{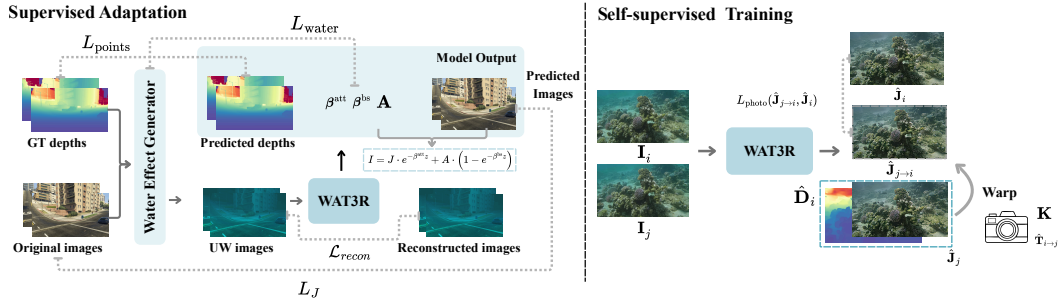}
  \caption{Overview of the proposed learning pipeline. \textbf{Left:} Supervised adaptation with a physical water effect generator. \textbf{Right:} Self-supervised training via cross-view photometric consistency.}
  \label{fig:training}
  \vspace{-15pt}
\end{figure}

To this end, we propose a framework that estimates scene geometry by explicitly leveraging UIFM process, as illustrated in \Cref{fig:pipeline}. Grounded in the physical imaging model (\Cref{sec:preliminary}), our method extends a feed-forward reconstruction pipeline to underwater environments. Specifically, the architecture (\Cref{sec:model_structure}) consists of three key components: a 3D reconstruction branch for dense geometry and camera pose estimation, a lightweight neural adaptation module for radiometric correction, and a water parameter estimation head for modeling global physical effects. For training, we first leverage physically grounded synthetic data for full supervision (\Cref{sec:synthetic}), and then adapt the model to real-world data through self-supervised multi-view consistency (\Cref{fig:training}). This design enables radiometric recovery to effectively enhance reconstruction accuracy and robustness.


\subsection{Preliminary}\label{sec:preliminary} \textbf{Underwater Image Formation Model}. It is tightly coupled with scene geometry through depth-dependent light attenuation and backscattering effects, which we leverage to enable self-supervised supervision of a feed-forward reconstruction model to underwater environments. Following~\cite{Akkaynak2018}, the observed intensity $I_c$ for a color channel $c \in \{R, G, B\}$ is:
\begin{equation}
I_c = J_c \cdot e^{-\beta_c^{\mathrm{att}} z} + B_c^\infty \cdot \big(1 - e^{-\beta_c^{\mathrm{bs}} z}\big),
\label{eq:uw_image_model}
\end{equation}
where $J_c$ is the true radiance, $z$ is the scene depth, and $\beta_c^{\mathrm{att}}, \beta_c^{\mathrm{bs}}, B_c^\infty$ denote the $c$-th components of the global parameters $\boldsymbol{\beta}^{\mathrm{att}}, \boldsymbol{\beta}^{\mathrm{bs}}, B^\infty \in \mathbb{R}^3$, respectively. The first term models direct transmission, decaying with depth and reducing brightness and contrast; the second captures backscattering. Their depth- and wavelength-dependence explains the severe color casts and contrast degradation in deeper regions.

Inspired by this model, our framework leverages the strong relationship between geometry and radiance in two ways. Methodologically, we first exploit \Cref{eq:uw_image_model} to construct a large-scale synthetic dataset with controllable degradation, where ground-truth depth and camera parameters enable fully supervised adaptation of our model under physically consistent underwater conditions. Architecturally, we then incorporate an auxiliary supervisory signal from the degradation adaptation module, allowing radiometric correction to provide complementary cues that further improve geometric estimation in underwater scenarios through self-supervised learning.


\textbf{Discussion}. ~\Cref{eq:uw_image_model} is a simplified and imperfect underwater imaging model. Thus, our method does not put ~\Cref{eq:uw_image_model} as a hard constraint in the geometry estimation. Instead, we apply it to construct a pseudo dataset for the domain adaptation. Then, we formulate a loss function based on the equation, which acts like a soft regularization to guide our model to improve the geometry estimation quality.

\subsection{Model Structure}\label{sec:model_structure}
To effectively model scene geometry under radiance variations as formulated in~\Cref{sec:preliminary}, we build our architecture on a shared feature extraction backbone that produces rich multi-scale representations for 3D reconstruction. In addition, we introduce a neural degradation adaptation module and a water parameter estimation head to model scene-specific radiometric effects. The degradation adaptation module predicts a degradation-aware clean image under varying underwater conditions, providing an auxiliary self-supervised signal that encourages degradation-invariant geometric features. Meanwhile, the water parameter estimation head captures scene-dependent optical properties of the water medium. Together, these components improve robustness to visual corruption and enable more accurate geometry estimation.

\textbf{3D Reconstruction Branch.}
Our 3D reconstruction branch follows the architecture proposed in~\cite{wang2025pi}. Specifically, the input RGB images are first processed by an encoder to produce patch tokens. These patch tokens, together with a set of register tokens, are then fed into a decoder to obtain refined representations, which are subsequently passed to downstream prediction branches. With one critical modification, we replace their original point prediction head with a Dense Prediction Transformer (DPT) head~\cite{Ranftl2021DPT}. This adjustment is designed to eliminate the grid-like artifacts typically introduced by the original upsampling MLP that relies on pixel-shuffling operations, thereby yielding smoother and more geometrically consistent depth predictions.

\textbf{Neural Degradation Adaptation Module.} This module predicts a degradation-aware clean image to provide self-supervised guidance under varying underwater conditions. It leverages intermediate encoder features $\mathbf{X}_i^l \in \mathbb{R}^{C_l \times H_l \times W_l}$, extracted from the $l$-th stage of the encoder for view $i$, which capture rich hierarchical image textures. These multi-scale features are first projected to a common dimensional space and then processed by an aggregator to fuse information across different scales:
\begin{equation}
\mathbf{F}_i = \mathcal{A} \Bigg( \sum_{l=1}^{L} \mathrm{Proj}(\mathbf{X}_i^l) \Bigg),
\end{equation}
where $\mathrm{Proj}(\cdot)$ denotes the projection operation, and $\mathcal{A}(\cdot)$ represents the aggregator module. The fused features $\mathbf{F}_i$ are subsequently passed through upsampling blocks to progressively recover spatial resolution, producing a residual map $\mathbf{R}_i$ that estimates the underlying degradation. The final clean image is obtained by adding this residual to the original input:
\begin{equation}
{\mathbf{J}}_i = \mathrm{clip}\big(\mathbf{I}_i + \mathbf{R}_i, 0, 1 \big).
\end{equation}
Predicting the degradation as a residual stabilizes the training process and forces the network to focus on modeling corruption patterns while effectively preserving the high-frequency fine details of the original images.

\textbf{Water Parameter Estimation Head.} To predict scene-level water parameters, this head utilizes the decoder's register tokens $\mathbf{r}_i$, which encapsulate the global context of the scene. The parameters are regressed through a lightweight Multi-Layer Perceptron (MLP), denoted as $f$:
\begin{equation}
\mathbf{w}_i = f(\mathbf{r}_i),
\end{equation}
where $\mathbf{w}_i = \{ \boldsymbol{\beta}^{\mathrm{att}}, \boldsymbol{\beta}^{\mathrm{bs}}, B^\infty \}$ represents the estimated global physical parameters. This architectural design explicitly decouples the estimation of global physical environment from the degradation adaptation process, enabling the model to effectively leverage both global context information.

\subsection{Training Strategy}

\begin{wrapfigure}{r}{0.45\textwidth} 
  \centering
  \includegraphics[width=\linewidth]{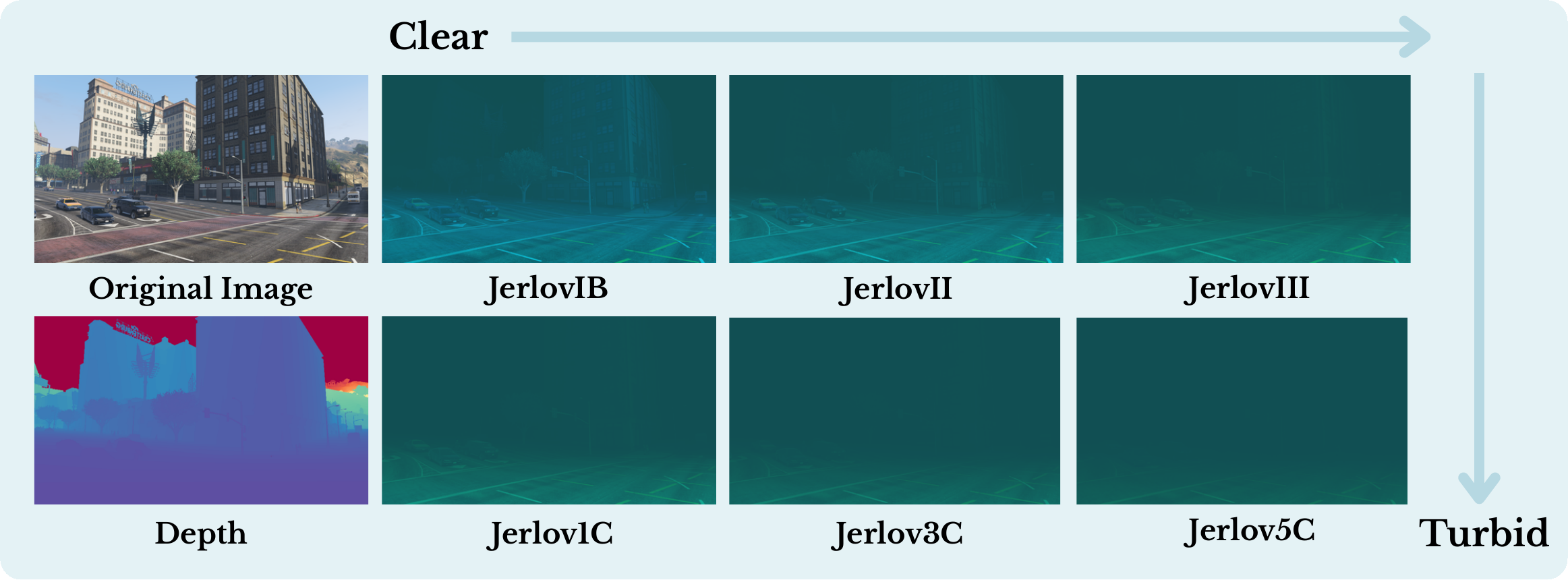} 
  \caption{Simulation of various underwater optical environments. Based on the clean in-air image and metric depth map (left), we synthesize physically plausible observations across different Jerlov water types.}
  \label{fig:water_types}
  \vspace{-8pt}
\end{wrapfigure}

\textbf{Synthetic Data Generation.}\label{sec:synthetic} To overcome the scarcity of real-world underwater datasets with high-quality geometric annotations, we leverage large-scale terrestrial RGB-D datasets that provide metrically accurate depth maps and camera poses to serve as full geometric supervision for underwater adaptation of pretrained feedforward models. Using the in-air images and their corresponding depth maps, we synthesize physically plausible underwater observations based on the underwater image formation model formulated in~\Cref{eq:uw_image_model}.

To ensure strict physical realism, the attenuation and backscattering coefficients ($\beta_c^{\mathrm{att}}$, $\beta_c^{\mathrm{bs}}$) are rigorously sampled from the measured inherent optical properties (IOPs) of Jerlov water types~\cite{Williamson2022}. Specifically, the attenuation and backscattering coefficients are derived from a comprehensive ocean optics database encompassing six representative water conditions, ranging from optically clear (Jerlov IB) to highly turbid (Jerlov 5C) waters.~\Cref{fig:water_types} illustrates the visual variations across these simulated water types. Furthermore, the global background light $B_c^\infty$ is estimated following the ULAP model~\cite{Song2018}, utilizing distinct background light samples from the UVEB dataset~\cite{Xie2024} to simulate diverse underwater illumination conditions.

\textbf{Supervised Adaptation on Synthetic Data.}
We adapt the network to the underwater domain in an end-to-end manner:
\begin{equation}
\mathcal{L} =
\lambda_J \mathcal{L}_J +
\lambda_{\mathrm{recon}} \mathcal{L}_{\mathrm{recon}} +
\lambda_{\mathrm{water}} \mathcal{L}_{\mathrm{water}} +
\lambda_{\mathrm{points}} \mathcal{L}_{\mathrm{points}} +
\lambda_{\mathrm{cam}} \mathcal{L}_{\mathrm{cam}}.
\end{equation}

Here, $\mathcal{L}_J$ is the degradation adaptation loss, which combines $L_1$ reconstruction, perceptual loss, and SSIM loss to ensure photometric and structural fidelity of the clean image. $\mathcal{L}_{\mathrm{recon}}$ enforces physical consistency by matching the input underwater image with the re-rendered image produced by the underwater imaging model. $\mathcal{L}_{\mathrm{water}}$ is an $L_2$ regression loss for supervising global water parameter estimation. For geometry learning, $\mathcal{L}_{\mathrm{points}}$ is a depth-weighted $L_1$ loss applied to pixel-aligned 3D point predictions after scale alignment, while $\mathcal{L}_{\mathrm{cam}}$ supervises relative camera poses using rotation distance and a Huber loss on translation error. Using synthetic data provides full supervision of all components, enabling the model pretrained on the terrestrial images to first adapt to underwater data. Then, we will further apply self-supervised losses on underwater data.

\textbf{Self-supervised Training on Real-world Data.} To bridge the domain gap between synthetic terrestrial training data and real underwater environments, WAT3R is subsequently adapted to unlabeled real-world underwater sequences via a self-supervised learning method, inspired by~\cite{Godardmonodepth2}. This utilizes the consistency between neighboring frames by warping the neighboring frames to each other according to depth estimation and then enforcing the consistency.

Specifically, the predicted clean images $\hat{\mathbf{J}}$ serve as a canonical, water-free scene representation that is strictly consistent across multiple views. For a given source view $i$ and a target view $j$, we utilize the predicted source depth $\hat{\mathbf{D}}_i$, the relative camera pose $\hat{\mathbf{T}}_{i \to j}$, and the camera intrinsics $\mathbf{K}$ to synthesize the corresponding clean image in the source coordinate frame via differentiable inverse warping:\begin{equation}\hat{\mathbf{J}}_{j \to i} = \mathrm{Warp}\big(\hat{\mathbf{J}}_j, \hat{\mathbf{D}}_i, \hat{\mathbf{T}}_{i \to j}, \mathbf{K}\big).\end{equation}The self-supervised objective enforces photometric consistency between the synthesized clean views and the predicted source clean views. To robustly handle occlusions and dynamic moving objects, we apply an auto-masking mechanism $\mathcal{M}_i$ to filter out unreliable regions:
\begin{equation}\mathcal{L}_{\mathrm{self}} = \mathcal{L}_{\mathrm{photo}}\big(\hat{\mathbf{J}}_{j \to i}, \hat{\mathbf{J}}_i\big) \odot \mathcal{M}_i.\end{equation}
Empirically, while the formation model (\Cref{eq:uw_image_model}) robustly constrains depth estimation, the domain shift from synthetic to real data disproportionately degrades camera pose prediction. This inaccuracy cascades into depth maps during view synthesis, causing unstable geometric reconstructions. Therefore, in the self-supervised training stage, we focus on optimizing the camera pose using the clean images. By minimizing the discrepancy between water-free representations, we effectively stabilize the pose estimation and ensure superior geometric consistency across real-world sequences.

%% file: Sec/4_experiment.tex
\section{Experiments}
\subsection{Multi-view Depth Estimation}
\textbf{Benchmarks and metrics.}
We evaluate our model on two underwater benchmarks: FLSea-Canyons~\cite{randall2023flsea} and SQUID~\cite{berman2020underwater}. Both datasets provide ground-truth depth annotations for underwater scenes with varying levels of turbidity and structural complexity. Following prior works~\cite{wang2025vggt, wang2025pi}, we report results under two alignment protocols: (1) \textit{scale-only} alignment and (2) \textit{scale-and-shift} alignment, ensuring fair comparison across different normalization settings. For quantitative evaluation, we adopt two standard depth estimation metrics: Absolute Relative Error (Abs Rel) and the accuracy under the threshold $\delta < 1.25$. Lower values of Abs Rel indicate better depth accuracy, while higher accuracy reflects more reliable pixel-wise depth prediction.

\begin{figure}[t]
  \centering
  \includegraphics[width=0.9\textwidth]{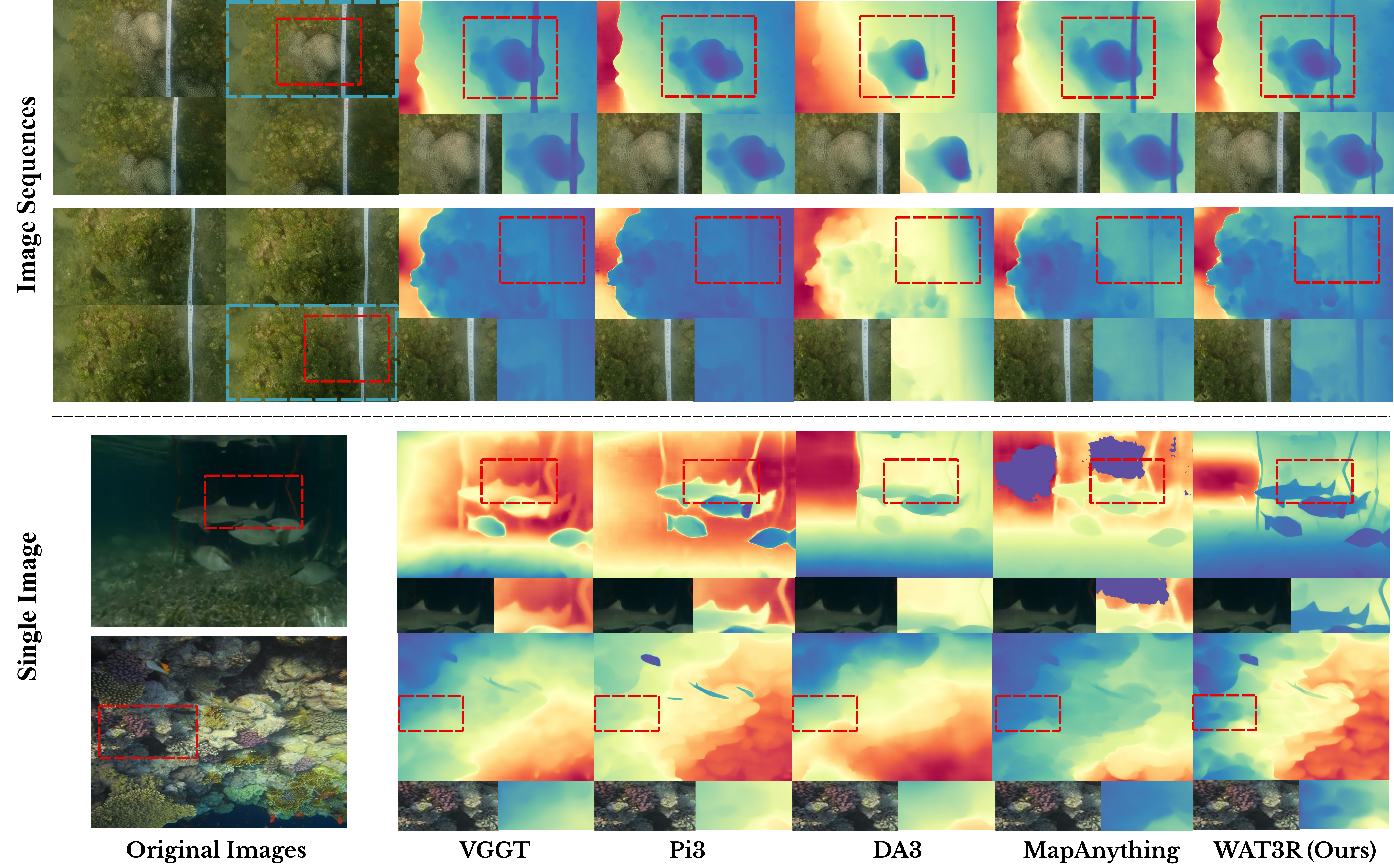}
  \caption{Qualitative comparison of depth estimation results on underwater scenes. Colors indicate relative depth from near (blue) to far (red).}
  \label{fig:depth-display}
  \vspace{-10pt}
\end{figure}
\begin{figure}[htbp]
  \centering
  \includegraphics[width=0.9\textwidth]{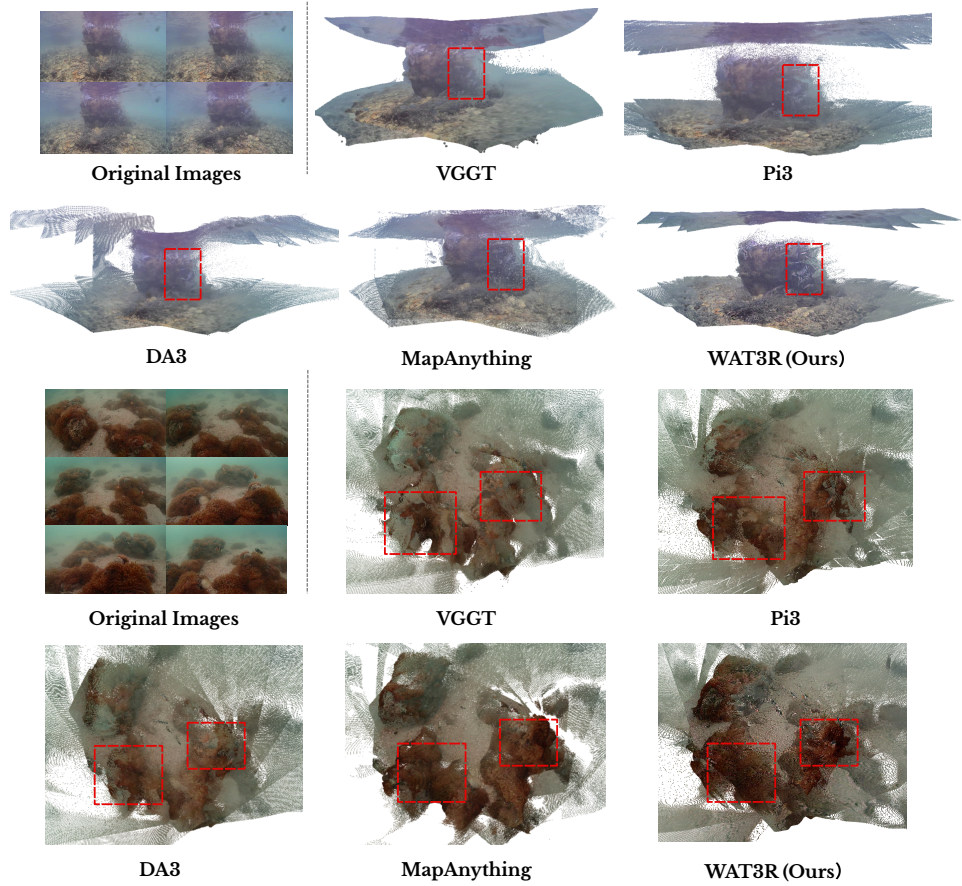}
  \caption{Qualitative comparison of 3D point cloud reconstruction on underwater scenes. Compared with existing methods, our proposed WAT3R achieves more complete geometry and finer details, especially in the regions highlighted by the red dashed boxes.}
  \label{fig:point-display}
\vspace{-15pt}
\end{figure}

\begin{table}[htbp]
  \centering
  \small
\caption{\textbf{Multi-view Depth Estimation on FLSea-Canyons~\cite{randall2023flsea} and SQUID~\cite{berman2020underwater} datasets.} We report error metrics (Abs Rel, $\delta < 1.25$) for both \textit{scale-only} and \textit{scale-and-shift} alignment. Best results are highlighted in \colorbox{bestblue}{darker blue} and second best are highlighted in \colorbox{secondblue}{light blue}.}
\label{tab:multi_view_depth_estimation}
\begin{adjustbox}{width=0.6\textwidth}
\begin{tabular}{l cc c cc}
\toprule
\multirow{2}{*}{\textbf{Method}} & \multicolumn{2}{c}{\textbf{FLSea-Canyons}} & & \multicolumn{2}{c}{\textbf{SQUID}} \\
\cmidrule{2-3} \cmidrule{5-6}
& Abs Rel $\downarrow$ & $\delta < 1.25 \uparrow$ & & Abs Rel $\downarrow$ & $\delta < 1.25 \uparrow$ \\
\midrule
\multicolumn{6}{l}{\textit{Alignment: Scale}} \\
\midrule
UDepth\cite{yu2022udepth}  & 0.3068 & 0.3918 & & 0.4152 & 0.2744 \\
TRUDepth \cite{TRUDepth} & \cellcolor{secondblue}{0.0728} & 0.9490 & & 0.2917 & 0.4764\\
VGGT   \cite{wang2025vggt}            & 0.1263 & 0.8712 & & 0.1632 & \cellcolor{secondblue}{0.8215} \\
Pi3   \cite{wang2025pi}      & 0.0891 & \cellcolor{secondblue}{0.9613} & & 0.1959 & 0.7740 \\
DA3 \cite{lin2025DA3}       & 0.0943 & 0.9345 & & \cellcolor{secondblue}{0.1570} & 0.7753\\
MapAnything \cite{keetha2026mapanything}    & 0.1251 & 0.8566 & & 0.1585 & 0.7880 \\
\midrule
\textbf{WAT3R} (Ours) & \cellcolor{bestblue}{0.0725}  & \cellcolor{bestblue}{0.9726} & & \cellcolor{bestblue}{0.1486} & \cellcolor{bestblue}{0.8501} \\
\midrule
\multicolumn{6}{l}{\textit{Alignment: Scale and Shift}} \\
\midrule
UDepth\cite{yu2022udepth}  & 0.3296 & 0.5277 & & 0.4555 & 0.4905 \\
TRUDepth \cite{TRUDepth} & 0.0698 & 0.9518 & & 0.2465 & 0.5395 \\
VGGT \cite{wang2025vggt}           & 0.1209 & 0.8700 & & 0.1501 & 0.8575 \\
Pi3 \cite{wang2025pi}        & \cellcolor{secondblue}{0.0819} & \cellcolor{secondblue}{0.9660} & & 0.1832 & \cellcolor{bestblue}{0.8994} \\
DA3  \cite{lin2025DA3} & 0.0929 & 0.9282 & & 0.1384 & 0.8356\\
MapAnything \cite{keetha2026mapanything}    & 0.1147 & 0.8757 & & \cellcolor{bestblue}{0.1332} & 0.8380 \\
\midrule
\textbf{WAT3R} (Ours) & \cellcolor{bestblue}{0.0693} & \cellcolor{bestblue}{0.9728} & & \cellcolor{secondblue}{0.1369} & \cellcolor{secondblue}{0.8964} \\
\bottomrule
\end{tabular}
\end{adjustbox}
\end{table}

\textbf{Comparison with existing methods.}
Quantitative comparisons are summarized in ~\Cref{tab:multi_view_depth_estimation}. Across both datasets and alignment settings, WAT3R consistently achieves the best or highly competitive performance. On FLSea-Canyons, our method obtains the lowest Abs Rel under both alignment strategies, while also achieving the highest $\delta < 1.25$ accuracy. On the SQUID dataset, WAT3R achieves the lowest Abs Rel and maintains competitive accuracy, demonstrating strong robustness and cross-dataset generalization.

\Cref{fig:depth-display} presents comparisons on representative underwater scenes in terms of 2D depth maps. Existing methods often suffer from blurred object boundaries or fragmented artifacts: MapAnything~\cite{keetha2026mapanything} exhibits obvious patch-wise discontinuities, while VGGT~\cite{wang2025vggt} and Pi3~\cite{wang2025pi} show depth bleeding around thin structures like fish and corals. In contrast, WAT3R preserves sharper contours and clearer foreground–background separation. Furthermore, as visualized in~\Cref{fig:point-display}, our method effectively avoids the geometric distortions and ``flying pixels'' common in DA3~\cite{lin2025DA3} and other baselines, maintaining superior structural coherence and realistic surface geometry. These qualitative improvements are consistent with the quantitative gains reported in \Cref{tab:multi_view_depth_estimation}.

\subsection{Monocular Depth Estimation}
\textbf{Benchmarks and metrics.} We evaluate monocular depth estimation on USOD10K~\cite{Lin2025usod10k} benchmark. It should be noted that the depth maps in the USOD10K~\cite{Lin2025usod10k} dataset are pseudo-depth maps generated by the DPT~\cite{Ranftl2021DPT} method, rather than true depth values measured by hardware sensors. As a result, they may contain inaccuracies compared with real physical depths, especially in complex underwater environments. Nevertheless, we use them as a geometric reference to compare the performance of different models. For the metrics, We continue to use two standard depth estimation metrics to evaluate our monocular approach: Absolute Relative Error (Abs Rel), and the accuracy under the threshold $\delta < 1.25$.

\noindent\textbf{Comparison with existing methods}. 
As shown in ~\Cref{tab:monocular_depth}, we compare WAT3R with prior methods on USOD10K. UDepth and TRUDepth are monocular depth models, while others use feedforward reconstruction. Despite this, WAT3R achieves the best Abs Rel (0.9046), demonstrating stronger global depth robustness and clear gains over UDepth and TRUDepth. Although DA3 slightly outperforms on $\delta < 1.25$, WAT3R maintains better overall balance across metrics, indicating more stable performance in complex underwater scenes.

\begin{table}[htbp]
\vspace{-10pt}
  \centering
  \small
  \begin{minipage}[c]{0.47\textwidth}
    \centering
    \caption{\textbf{Monocular Depth Estimation on USOD10K~\cite{Lin2025usod10k} dataset.} We report error metrics (Abs Rel, $\delta < 1.25$). Best results are highlighted in \colorbox{bestblue}{darker blue} and second best are highlighted in \colorbox{secondblue}{light blue}.}
    \label{tab:monocular_depth}
    \scalebox{0.8}{
    \begin{tabular}{lcc} 
      \toprule
      \textbf{Method} & Abs Rel $\downarrow$ & $\delta < 1.25$ $\uparrow$ \\
      \midrule
      UDepth ~\cite{yu2022udepth} & 0.9250 & 0.3310 \\
      TRUDepth ~\cite{TRUDepth} & 1.3036 & 0.4413 \\
      VGGT ~\cite{wang2025vggt} & 1.2612 & 0.4700 \\
      Pi3  ~\cite{wang2025pi} & \cellcolor{secondblue}{0.9851} & 0.4720 \\
      DA3  ~\cite{lin2025DA3} & 1.1322 & \cellcolor{bestblue}{0.5330} \\
      MapAnything~\cite{keetha2026mapanything}  & 1.0013 & 0.4702 \\
      \midrule
      \textbf{WAT3R (Ours)} & \cellcolor{bestblue}{0.9046} & \cellcolor{secondblue}{0.4916} \\
      \bottomrule
    \end{tabular}}
  \end{minipage}
  \hfill 
  \begin{minipage}[c]{0.49\textwidth}
    \centering
    \caption{\textbf{Camera pose estimation on our self-collected real underwater dataset.} We report standard evaluation metrics including ATE, RPE$_{trans}$ and RPE$_{rot}$. \colorbox{bestblue}{Darker blue} and \colorbox{secondblue}{light blue} indicate the best and second-best results.}
    \label{tab:self_collected}
    \resizebox{1\linewidth}{!}{ 
      \begin{tabular}{lccc}
        \toprule
        Model & ATE $\downarrow$ & RPE$_{trans}$ $\downarrow$ & RPE$_{rot}$ $\downarrow$ \\
        \midrule
        VGGT~\cite{wang2025vggt} & 0.2473 & 0.0846 & 0.4181 \\
        Pi3~\cite{wang2025pi} & 0.2353 & \cellcolor{bestblue}{0.0680} & \cellcolor{secondblue}{0.3525} \\
        DA3~\cite{lin2025DA3} & \cellcolor{bestblue}{0.2025} & 0.0856 & 0.4133 \\
        MapAnything~\cite{keetha2026mapanything} & 0.2965 & 0.1148 & 0.6570 \\
        \midrule
        \textbf{WAT3R (Ours)} & \cellcolor{secondblue}{0.2321} & \cellcolor{secondblue}{0.0721} & \cellcolor{bestblue}{0.2777} \\
        \bottomrule
      \end{tabular}
    }
  \end{minipage}
\end{table}

\subsection{Camera Pose Estimation}
\textbf{Benchmarks and metrics.} We evaluate camera pose estimation on a self-collected real-world underwater dataset captured in natural marine environments. The dataset contains challenging motion patterns induced by water currents, dynamic illumination changes, and feature-sparse regions, making accurate pose recovery particularly difficult. For quantitative evaluation, we adopt standard trajectory evaluation metrics, including Absolute Trajectory Error (ATE), Relative Translation Error (RPE$_{trans}$), and Relative Rotation Error (RPE$_{rot}$). Lower values indicate better pose accuracy and trajectory consistency.

\textbf{Comparison with existing methods.}~\Cref{tab:self_collected} summarizes the quantitative evaluation of camera pose estimation across different models. While our \textsc{WAT3R} achieves a significant breakthrough in rotational stability, reducing $\text{RPE}_{rot}$ by $21.2\%$ compared to Pi3~\cite{wang2025pi}, we observe a marginal decrease in $\text{ATE}$ and $\text{RPE}_{trans}$. This phenomenon is primarily attributed to the optimization priority shift during self-supervised adaptation on real-world underwater sequences. Specifically, the self-supervised objective tends to prioritize capturing complex angular dynamics caused by water currents to minimize photometric error, which can introduce subtle scale drift in feature-sparse underwater environments.

\subsection{Ablation study}
\textbf{Ablation on the different components.} To evaluate the contribution of each component in our proposed framework, we conduct comprehensive ablation studies.  Specifically, the different configurations are as follows: (1) without neural degradation adaptation module (w/o NDAM): we predict depth and camera poses directly from degraded underwater images and supervised solely by ground-truth. (2)
without reconstruction loss (w/o $\mathcal{L}_{\mathrm{recon}}$): in this setting, we employ a two-stage approach: the images are first fed into a neural degradation adaptation module, and gradients are not allowed to flow back, which prevents joint optimization. (3) with reconstruction loss (ours): the inclusion of $\mathcal{L}_{\mathrm{recon}}$ allows the degradation adaptation process to provide explicit geometric constraints, effectively guiding the model to learn physics-consistent features for reconstruction

\begin{table}[htbp]
\vspace{-10pt}
  \centering
  \begin{minipage}[t][4cm][t]{0.47\textwidth}
    \centering
    \caption{\textbf{Ablation study on FLsea-canyons datasets.} \colorbox{bestblue}{Darker blue} indicates the best performance and \colorbox{secondblue}{light blue} indicates the second best performance.}
    \label{tab:ablation_results}
    \small
    \setlength{\tabcolsep}{1.2pt}
    \begin{tabular}{l | cc}
      \toprule
      Method & Abs Rel $\downarrow$ & $\delta < 1.25$ $\uparrow$ \\
      \midrule
      (1) w/o NDAM & 0.0773 & 0.9526 \\
      (2) w/o $\mathcal{L}_{\mathrm{recon}}$ & \cellcolor{secondblue}{0.0704} & \cellcolor{secondblue}{0.9722} \\
      \midrule
      (3) \textbf{Ours} & \cellcolor{bestblue}{0.0692} & \cellcolor{bestblue}{0.9728} \\
      \bottomrule
    \end{tabular}
  \end{minipage}
  \hfill 
  \begin{minipage}[t][4cm][t]{0.47\textwidth}
    \centering
    \caption{\textbf{Ablation on self-supervised adaptation and domain gap analysis.} We compare camera pose estimation with and without self-supervised fine-tuning. \colorbox{bestblue}{Darker blue} indicates the best performance.}
    \label{tab:self_supervised}
    \small
    \setlength{\tabcolsep}{1.5pt}
    \vspace{5pt}
    \begin{tabular}{lccc}
      \toprule
      Setting & ATE $\downarrow$ & RPE$_{trans}$ $\downarrow$ & RPE$_{rot}$ $\downarrow$ \\
      \midrule
      w/o SSA & 0.2332 & 0.0784 & 0.3982 \\
      w/ SSA  & \cellcolor{bestblue}{0.2321} & \cellcolor{bestblue}{0.0721} & \cellcolor{bestblue}{0.2777} \\
      \bottomrule
    \end{tabular}
  \end{minipage}
\end{table}

\textbf{Ablation on self-supervised adaptation (SSA) and domain gap analysis.}
We further evaluate the adaptability of our framework to real underwater datasets, where ground-truth clean images are unavailable. In this setting, the pretrained model is fine-tuned using only self-supervised signals, such as multi-view photometric warping loss.~\Cref{tab:self_supervised} shows that self-supervised training further improves rotation and translation accuracy over the pretrained model, demonstrating the framework’s ability to adapt to new underwater environments using only geometric and photometric cues.

\textbf{Observation.}
Beyond the reconstruction task, we observe that our adaptation module inherently exhibits strong restoration capability, despite not being explicitly designed for this purpose. The reconstructed images exhibit high visual fidelity, suggesting that the module implicitly prioritizes the underlying physics of underwater light degradation. Due to space constraints, we provide extensive qualitive and quantative analysis in the supplementary materials.

%% file: Sec/5_conclusion.tex
\section{Conclusion}
\vspace{-5pt}
In this paper, we introduce  \textbf{WAT3R}, a feed-forward framework for underwater 3D reconstruction that explicitly models the geometry-dependent characteristics of underwater image formation. By leveraging a neural degradation adaptation module with a reconstruction branch through a physics-guided underwater image formation model, WAT3R leverages it as an auxiliary supervisory signal to enhance 3D reconstruction accuracy. Unlike decoupled pipelines or optimization-based neural rendering approaches, our method enables physically consistent reconstruction in a scene-agnostic and computationally efficient manner. Furthermore, our hybrid training strategy combining large-scale synthetic supervision with self-supervised adaptation on real-world data significantly improves robustness across diverse underwater conditions. We believe this work represents a meaningful step toward physically grounded underwater scene reconstruction.

%% file: Sec/X_suppl.tex
\clearpage
\counterwithin{table}{section}
\counterwithin{figure}{section}
\section{Overview}
\label{sec:overview}

This supplementary material provides additional details for our WAT3R framework. 
Section \ref{sec:implementation} introduces the details of its implementation.
Section \ref{sec:Underwater_Image_Restoration} presents additional qualitative and quantitative analysis of underwater image restoration.
Section \ref{sec:synthetic_data_generation} describes details of the synthetic data generation pipeline based on measured inherent optical properties (IOPs) and Jerlov water types.
Section \ref{sec:camera_pose_estimation} presents our real underwater dataset and evaluates our pose estimation method against visual odometry baselines. 
Finally, Section \ref{sec:limitation} discusses current limitations in very turbid water and outlines possible future improvements.

\section{Implementation Details}
\label{sec:implementation}
Our model is trained in two stages. In the first stage, we focus on supervised adaptation to optimize the base geometric model, leveraging datasets such as VKITTI~\cite{cabon2020vkitti2}, Tartanair~\cite{tartanair2020iros}, and MVS-Synth~\cite{Huang2018DeepMVS} which provide high-quality ground-truth supervision. In the second stage, we conduct self-supervised training on the FLSea-RedSea~\cite{randall2023flsea} and DRUVA~\cite{Varghese_2023_druva} datasets to adapt the framework to complex real-world underwater conditions. We adopt the AdamW optimizer with a base learning rate of $5 \times 10^{-6}$ and a weight decay of 0.05 and for training efficiency, we use dynamic resolution scaling with a pixel count range of [100,000, 255,000] and a patch size of 14. All experiments are conducted using eight NVIDIA RTX 4090 GPUs, with Pi3~\cite{wang2025pi} serving as the backbone initialization. 

\subsection{Neural Degradation Adaptation Module}
It is an auxiliary module designed to mitigate underwater degradation by aggregating multi-scale latent features and enforcing spatial consistency. 

\paragraph{Feature Projection and Aggregation}
Given a set of intermediate feature maps $\{\mathbf{X}_i^l\}_{l \in \mathcal{L}}$ extracted from the $l$-th layer of the DINOv2 encoder for view $i$ (where $\mathcal{L}$ denotes the selected layer indices), we first project each feature map into a common embedding space:
\begin{equation}
    \hat{\mathbf{X}}_i^l = \mathrm{Proj}(\mathbf{X}_i^l) = \mathrm{Conv}_{1 \times 1}^{(l)}(\mathbf{X}_i^l) \in \mathbb{R}^{d_{proj} \times h \times w},
\end{equation}
where $h=H/14$ and $w=W/14$ represent the patch-level resolution. These multi-scale features are aggregated and fused to form the initial latent representation $\mathbf{F}_i$:
\begin{equation}
    \mathbf{F}_i = \mathcal{A} \Bigg( \sum_{l \in \mathcal{L}} \hat{\mathbf{X}}_i^l \Bigg) \in \mathbb{R}^{2d_{proj} \times h \times w},
\end{equation}
where $\mathcal{A}(\cdot)$ denotes the aggregator module consisting of a $1 \times 1$ convolution, GroupNorm, and GELU activation.

\paragraph{Iterative Conditional Upsampling}
To progressively recover high-frequency spatial details, we employ a series of $K=4$ conditional upsampling blocks $\Phi_k$. Let $\mathbf{x}_k$ be the feature map at stage $k$, with the initial state $\mathbf{x}_0 = \mathbf{F}_i$. At each stage, the feature is concatenated with a normalized UV coordinate grid $\mathbf{G}_k$ and the downsampled original image $\mathbf{I}_i^{(k)}$ to provide spatial and appearance priors:
\begin{equation}
    \mathbf{z}_k = \mathrm{Concat}(\mathbf{x}_{k-1}, \mathbf{G}_k, \mathbf{I}_i^{(k)}) \in \mathbb{R}^{(C_k + 2 + 3) \times H_k \times W_k}.
\end{equation}
The latent feature is then updated and upsampled through the block $\Phi_k$:
\begin{equation}
    \mathbf{x}_k = \Phi_k(\mathbf{z}_k) = \mathrm{Block}(\mathrm{Upsample}_{2\times}(\mathrm{Block}(\mathbf{z}_k))),
\end{equation}
where each $\mathrm{Block}(\cdot)$ denotes a sequence of $3 \times 3$ convolution, GroupNorm, and GELU activation.

\paragraph{Residual Reconstruction}
The final residual map $\mathbf{R}_i$, which estimates the underlying degradation pattern, is predicted at the original resolution $H \times W$ by fusing the final upsampled features with full-resolution coordinates and the input image:
\begin{equation}
    \mathbf{R}_i = \mathrm{Conv}_{out}(\mathrm{Concat}(\mathrm{Interp}(\mathbf{x}_K), \mathbf{G}_{full}, \mathbf{I}_i)).
\end{equation}
The final restored clean image $\hat{\mathbf{J}}_i$ is obtained by:
\begin{equation}
    \hat{\mathbf{J}}_i = \mathrm{clip}(\mathbf{I}_i + \mathbf{R}_i, 0, 1).
\end{equation}
By modeling the degradation as a residual $\mathbf{R}_i$, the network effectively preserves the fine textures of the original scene while stabilizing the training process through the auxiliary supervision of image reconstruction.

\subsection{Supervised Adaptation}
The training objective is defined as
\begin{equation}
\mathcal{L} =
\lambda_J \mathcal{L}_J +
\lambda_{\mathrm{recon}} \mathcal{L}_{\mathrm{recon}} +
\lambda_{\mathrm{water}} \mathcal{L}_{\mathrm{water}} +
\lambda_{\mathrm{points}} \mathcal{L}_{\mathrm{points}} +
\lambda_{\mathrm{cam}} \mathcal{L}_{\mathrm{cam}},
\end{equation}
where we set $\lambda_J = 0.5$, $\lambda_{\mathrm{recon}} = 0.3$, $\lambda_{\mathrm{points}} = 1.0$, $\lambda_{\mathrm{cam}} = 0.2$, and $\lambda_{\mathrm{water}} = 0.3$.

The restoration supervision loss is defined as
\begin{equation}
\mathcal{L}_J =
\lambda_{L1} \mathcal{L}_{L1} +
\lambda_{perc} \mathcal{L}_{perc} +
\lambda_{ssim} \mathcal{L}_{ssim}
\end{equation}

which enforces pixel fidelity, perceptual similarity and structural consistency between the prediction and the ground truth. We set $\lambda_{L1} = 0.8$, $\lambda_{perc} = 0.1$, and $\lambda_{ssim} = 0.2$. 

\noindent
The water parameter supervision is defined using the mean squared error:
\begin{equation}
\mathcal{L}_{\mathrm{water}} =
\left\| \mathbf{w}_{pred} - \mathbf{w}_{gt} \right\|_2^2 .
\end{equation}

The additional geometric losses follow~\cite{wang2025pi} and the network predicts a pixel-aligned 3D point map $\hat{\mathbf{X}}_i$ for each input image. 
To resolve monocular scale ambiguity, the predicted points are aligned to the ground truth using a global scale factor $\hat{s}$. 
The loss is defined as a depth-weighted L1 distance:
\begin{equation}
L_{\mathrm{points}} =
\sum_{i=1}^{N}\sum_{j=1}^{H\times W}
\frac{1}{z_{i,j}}
\left\| \hat{s}\hat{\mathbf{x}}_{i,j}-\mathbf{x}_{i,j} \right\|_1 .
\end{equation}

The network also predicts camera poses $\hat{\mathbf{T}}_i$. 
Using the same scale factor $\hat{s}$, we supervise the relative poses between views via rotation and translation losses:
\begin{equation}
L_{\mathrm{cam}} =
\frac{1}{N(N-1)}\sum_{i\neq j}
\bigl(L_{\mathrm{rot}}(i,j)+\lambda L_{\mathrm{trans}}(i,j)\bigr),
\end{equation}
where $L_{\mathrm{rot}}$ measures rotation distance and $L_{\mathrm{trans}}$ is a Huber loss on relative translation.

\subsection{Unsupervised Training}
To robustly handle occlusions and stationary pixels, we adopt a minimum reprojection loss and auto-masking strategy inspired by~\cite{Godardmonodepth2}. For each source view $i$ and target view $j$, the photometric loss for the clean (restored) image is computed as:\begin{equation}L_{\text{photo}} =\min_{j \neq i} \Big(\lambda_{ssim} \cdot \mathrm{SSIM}(\hat{I}_{j \to i}, \hat{I}i)
+
\lambda_{L1} \cdot \mathrm{L_1}(\hat{I}{j \to i}, \hat{I}_i)\Big),\end{equation}where $\hat{I}_{j \to i}$ is the synthesized image in the source coordinate frame, $\lambda_{ssim}=0.85$ and $\lambda_{L1}=0.15$. To ignore pixels that violate the motion assumption (e.g., stationary objects), we compute the identity reprojection loss:\begin{equation}L_{\text{id}} =\min_{j \neq i} \mathrm{L_1}(\hat{I}_{j}, \hat{I}_i),\end{equation}and construct the auto-mask:\begin{equation}M_i = \mathbf{1}\Big(L_{\text{photo}} < L_{\text{id}}\Big),\end{equation}where $\mathbf{1}(\cdot)$ is the indicator function. The final self-supervised loss $L_{\text{self}}$ is defined as the masked photometric loss:\begin{equation}L_{\text{self}} = M_i \odot L_{\text{photo}}.\end{equation}

\section{Underwater Image Restoration}
\label{sec:Underwater_Image_Restoration}
\noindent\textbf{Benchmarks and metrics.}
We conduct a comprehensive evaluation on both synthetic and real-world underwater benchmarks. 
For quantitative full-reference evaluation, we adopt the synthetic underwater dataset introduced by~\cite{Nathan2024osmosis}, constructed from NYUv2~\cite{Silbermannyuv2} indoor RGB-D images with high-quality ground-truth depth. 
For real-world evaluation, we further use two widely adopted benchmarks, USOD10k~\cite{Lin2025usod10k} and UIEB~\cite{li2019underwater}, which contain diverse underwater scenes with varying levels of color distortion, haze, and contrast degradation.

For synthetic data where ground-truth clean images are available, we follow standard full-resolution evaluation protocols and report full-reference metrics, including Peak Signal-to-Noise Ratio (PSNR) and Structural Similarity Index (SSIM). 
For real-world datasets without reference images, we adopt two non-reference image quality metrics, UIQM~\cite{Karen2016uiqm} and MUSIQ~\cite{ke2021musiq}, following~\cite{wu2025slurpp}. Higher scores indicate better perceptual quality.

\begin{figure}[htbp]
  \centering
  \includegraphics[width=\textwidth]{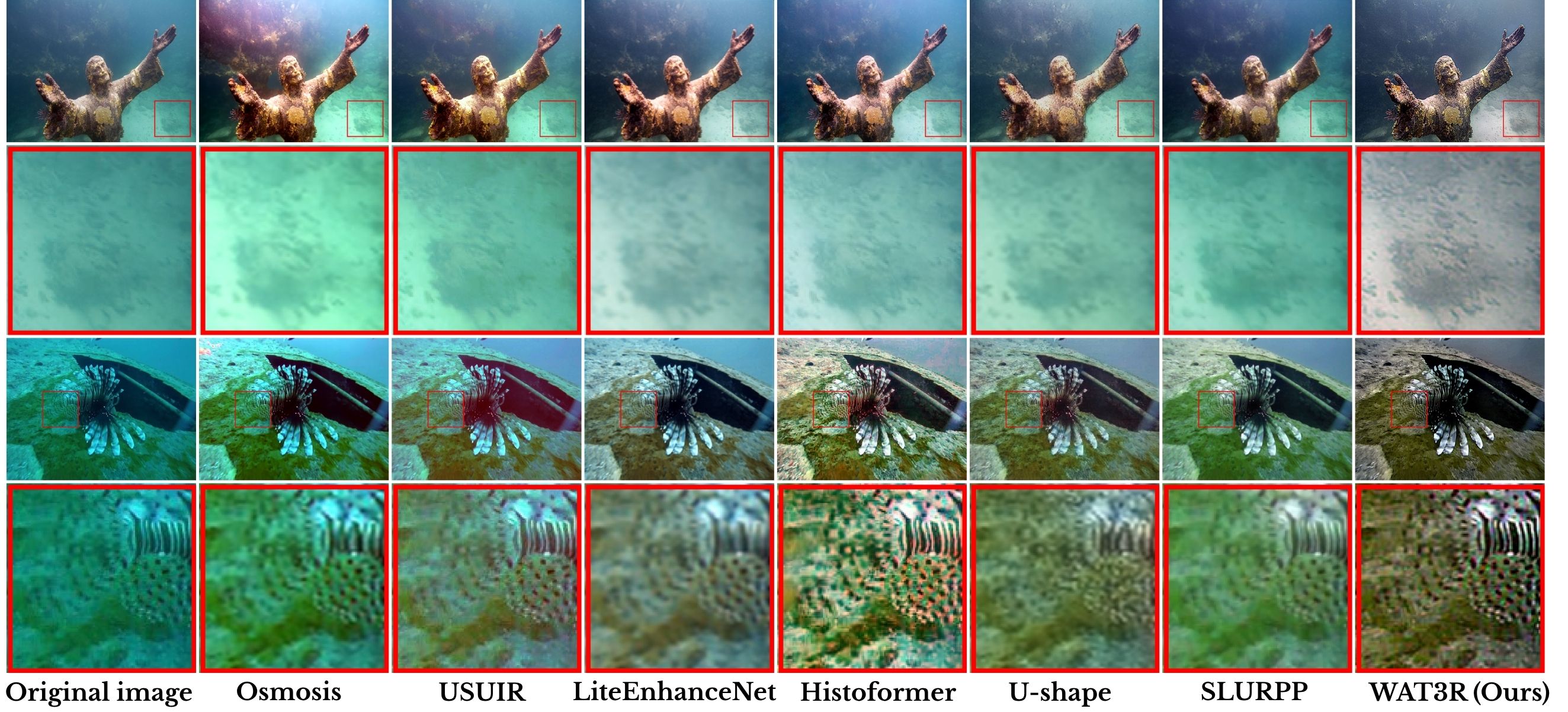}
  \caption{Qualitative comparison of underwater image restoration methods. From left to right, the visual quality and color restoration improve, with our method (rightmost) achieving the most natural results.}
  \label{fig:UIE-display}
\end{figure}

\begin{table}[htbp]
\centering
\caption{\textbf{Quantitative results on synthetic~\cite{Nathan2024osmosis}, USOD10k~\cite{Lin2025usod10k}, and UIEB~\cite{li2019underwater} datasets.} \colorbox{bestblue}{Darker blue} indicates the best performance, and \colorbox{secondblue}{light blue}indicates the second-best.}
\label{tab:restoration_metrics}
\small 
\begin{tabular}{lcccccc}  
\toprule
\multirow{2}{*}{Method} & \multicolumn{2}{c}{Synthetic Dataset} & \multicolumn{2}{c}{USOD10k} & \multicolumn{2}{c}{UIEB} \\ \cmidrule(lr){2-3} \cmidrule(lr){4-5} \cmidrule(lr){6-7}
 & PSNR $\uparrow$ & SSIM $\uparrow$ & UIQM $\uparrow$ & MUSIQ $\uparrow$ & UIQM $\uparrow$ & MUSIQ $\uparrow$ \\ \midrule
Histoformer~\cite{Peng2025Histoformer} & 16.15 & 0.773 & \cellcolor{bestblue}{2.645} & \cellcolor{secondblue}{42.23} & \cellcolor{secondblue}{2.908} & \cellcolor{bestblue}{52.34} \\
USUIR~\cite{Fu2022USUIR} & 16.76 & 0.801 & \cellcolor{secondblue}{2.609} & 42.18 & \cellcolor{bestblue}{3.039} & 51.21 \\
U-Shape~\cite{Peng2023U-shape} & 17.56 & 0.746 & 2.435 & 37.38 & 2.768 & 40.19 \\
LiteEnhanceNet~\cite{Zhang2024LiteEnhanceNet} & 17.88 & 0.835 & 2.377 & 36.33 & 2.641 & 37.41 \\
Osmosis~\cite{Nathan2024osmosis} & 22.12 & 0.890 & 2.397 & 38.68 & 2.659 & 41.30 \\
SLURPP~\cite{wu2025slurpp} & \cellcolor{secondblue}{24.89} & \cellcolor{bestblue}{0.922} & 2.280 & 40.67 & 2.751 & 45.82 \\ \midrule
\textbf{WAT3R (Ours)} &\cellcolor{bestblue}{26.81} & \cellcolor{secondblue}{0.917} & 2.343 & \cellcolor{bestblue}{44.02} & 2.727 & \cellcolor{secondblue}{52.29} \\ \bottomrule
\end{tabular}
\end{table}

\noindent\textbf{Comparison with existing methods.}
As shown in ~\Cref{tab:restoration_metrics}, WAT3R consistently outperforms existing methods on synthetic data, achieving the highest PSNR of 26.81\,dB while maintaining a competitive SSIM of 0.917. This demonstrates its strong capability in preserving pixel-level fidelity and recovering structural details under simulated underwater degradations. On real-world benchmarks, WAT3R achieves competitive performance compared to state-of-the-art specialized models such as Histoformer~\cite{Peng2025Histoformer}, notably attaining the highest MUSIQ scores across both datasets. This indicates superior perceptual quality and strong cross-task generalization, even without a dedicated restoration-specific design. Regarding UIQM, we observe that it does not always align with human perception, often favoring over-saturated or over-sharpened results. As illustrated in ~\Cref{fig:UIE-display}, WAT3R produces consistently more visually pleasing results than Histoformer despite lower UIQM scores. 

Overall, the combination of leading quantitative performance on synthetic data, superior perceptual quality on real-world benchmarks, and an efficient feedforward architecture demonstrates that WAT3R achieves a strong balance between fidelity, perceptual quality, and computational efficiency, outperforming iterative baselines~\cite{Nathan2024osmosis, wu2025slurpp}.

\begin{figure}[t]
  \centering
  \includegraphics[width=0.8\textwidth]{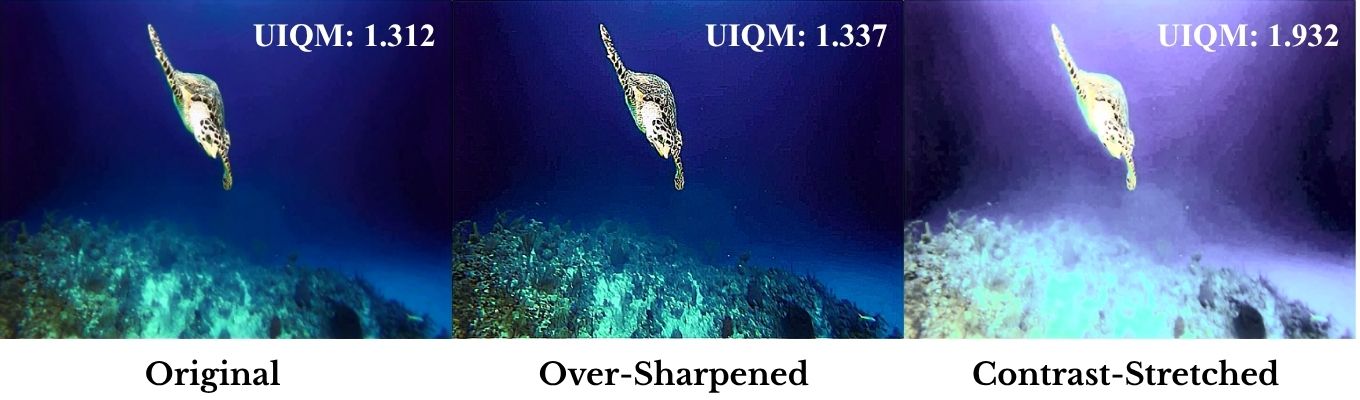}
  \caption{UIQM robustness analysis. Extreme enhancements artificially inflate UIQM scores despite introducing severe visual artifacts like posterization and halos.}
  \label{fig:uiqm}
\end{figure}

\textbf{UIQM Metric Analysis.} While UIQM is a widely used non-reference metric for underwater image quality, our experiments reveal a significant discrepancy between its numerical scores and actual perceptual quality. Theoretically, UIQM is defined as a linear combination of three attribute-specific components:
\begin{equation}
\text{UIQM} = c_1 \cdot \text{UICM} + c_2 \cdot \text{UISM} + c_3 \cdot \text{UIConM}
\end{equation}
where $c_1=0.0282$, $c_2=0.2953$, and $c_3=3.5753$
As demonstrated in ~\Cref{fig:uiqm}, the metric exhibits a pathological bias toward high-frequency noise and extreme pixel distributions. For instance, the 'Contrast-Stretched' version achieves the highest UIQM score (1.932) despite suffering from severe posterization and loss of natural texture. This is primarily because the exaggerated dynamic range heavily inflates the $\text{UIConM}$ component, which carries the largest weight coefficient ($c_3$). Similarly, 'Over-Sharpened' artifacts result in a higher score (1.337) than the original, even though they introduce distracting halos. These results suggest that UIQM's reliance on statistical color and gradient distributions makes it an unreliable proxy for evaluating restoration performance, as it tends to ``reward'' unnatural enhancements that are visually unappealing to human observers.

\section{Synthetic Data Generation}
\label{sec:synthetic_data_generation}
The synthetic data generation pipeline relies on the analysis of measured inherent optical properties (IOPs) of Jerlov water types~\cite{Williamson2022}. Specifically attenuation $a$ and backscattering coefficients $b$ for standard Jerlov water types using the World-wide Ocean Optics Database (WOOD). The process involves rigorous data cleaning and filtering to focus on the top 10 meters of the water column, followed by grouping over 13.5 million data points into "campaigns" based on spatial and temporal proximity. Each campaign is assigned a Jerlov water type by comparing its measured downwelling diffuse attenuation coefficient to standard Jerlov spectral profiles using a minimum Mean Absolute Percentage Error (MAPE) fitting method. 

Although Jerlov defined ten water types, this methodology focuses on six robust types (IB, II, III, 1C, 3C, and 5C) because they provided sufficient synchronized measurements for $a$ and $b$ to meet a statistical threshold of at least five independent entries. To create continuous spectral data from 300 to 800~nm, the discrete measured points were fitted to established bio-optical models that optimize parameters for chlorophyll and particulate matter concentrations. By sampling these validated profiles across the Red (630--745~nm), Green (500--575~nm), and Blue (460--495~nm) wavelength at a 10~nm step, the pipeline ensures that the synthetic dataset is grounded in physical reality rather than purely theoretical assumptions.

This approach provides significant advantages for underwater imaging by offering physically consistent parameters that correct discrepancies found in previous models, such as those that predicted too little absorption or too much scatter. While the source data has certain geographic limitations---with most campaigns located between 20$^{\circ}$ and 45$^{\circ}$ North---it represents the first comprehensive effort to link Jerlov classifications to large-scale measured inherent optical property (IOP) data. By utilizing these experimentally validated coefficients, the synthetic dataset achieves higher fidelity in simulating the distinct optical behaviors of diverse marine environments ranging from clear open oceans to turbid coastal waters.

A further consideration is the applicability of the underlying image formation assumptions under spatially varying water conditions. The proposed pipeline assumes that inherent optical properties (IOPs), namely the attenuation $a(\lambda)$ and backscattering $b(\lambda)$ coefficients, are locally homogeneous within each campaign-derived water type. This approximation is justified by the spatial-temporal aggregation used in constructing campaigns, which effectively averages fine-scale fluctuations while preserving the dominant optical characteristics associated with each Jerlov class.

In natural ocean environments, IOPs may still exhibit localized heterogeneity due to processes such as plankton patchiness, sediment resuspension, or vertical stratification. In such cases, the effective optical response may deviate from a strictly homogeneous medium, introducing additional variability in the observed image formation process beyond a simple depth-dependent attenuation model. Nevertheless, these effects are partially mitigated in practice by the fact that the dominant image degradation mechanisms in underwater imaging are still governed by bulk attenuation and backscattering behavior.

Importantly, our evaluation is conducted on multiple real-world underwater datasets that inherently include such non-uniform and uncontrolled environmental conditions. The observed improvements in 3D reconstruction quality on these datasets indicate that the proposed synthetic data generation pipeline remains effective even under realistic water variability, and provides meaningful generalization beyond the simplified homogeneous assumption used in data construction.

\section{Camera Pose Estimation}
\label{sec:camera_pose_estimation}
\subsection{Self-Collected Real Underwater Dataset}
\begin{wrapfigure}{r}{0.45\textwidth} 
  \centering
  \includegraphics[width=0.4\textwidth]{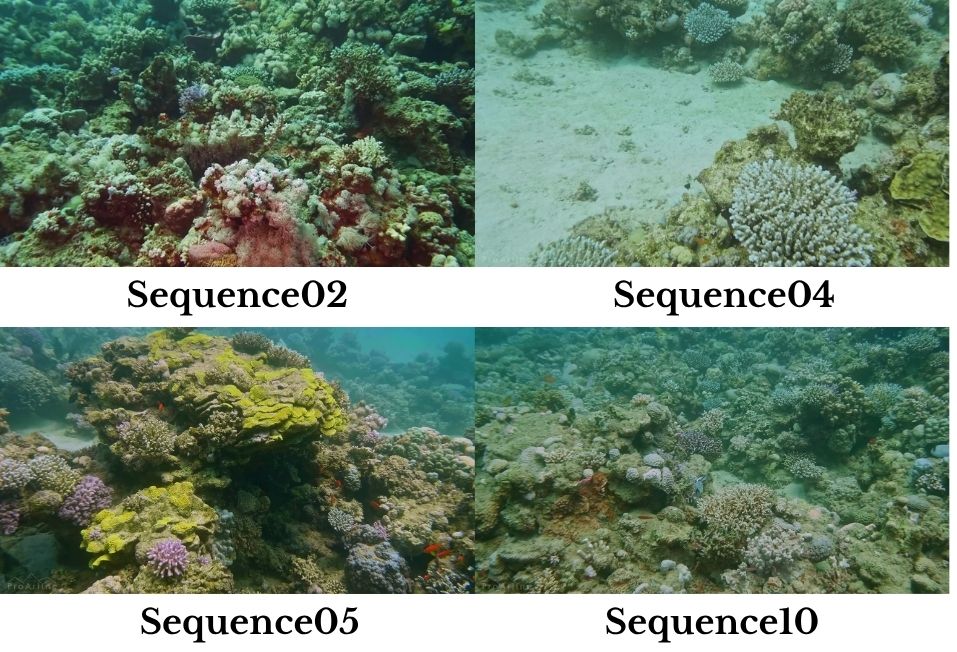}
  \caption{Representative sequences from our underwater dataset. Our data captures diverse benthic terrains, from complex coral reefs to sandy substrates.}
  \label{fig:cam-display}
\end{wrapfigure}

We construct a real-world underwater dataset composed of 10 video sequences, all sourced from publicly available underwater videos. Each sequence captures diverse underwater scenes with varying lighting, turbidity, and camera motion as illustrated in ~\Cref{fig:cam-display}.  

For each sequence, we first extract frames and use COLMAP~\cite{schoenberger2016mvs, schoenberger2016sfm} to obtain dense camera poses and we sample frames with a stride of 2. This dataset provides a realistic benchmark for assessing the performance of our camera pose estimation framework in real-world underwater conditions.
\subsection{Comparison with Visual Odometry}
We evaluate the camera pose estimation performance of our method by comparing it with a representative visual odometry (VO) baseline, ORB-SLAM3~\cite{ORBSLAM3}. Although ORB-SLAM3~\cite{ORBSLAM3} is designed as a full SLAM system, it can also operate in a visual odometry setting when global loop closure and map reuse are not considered. In this configuration, the system estimates the camera trajectory incrementally based on feature extraction, feature matching, and motion optimization between consecutive frames.

~\Cref{tab:slam_comparison} reports the quantitative comparison using standard trajectory evaluation metrics, including Absolute Trajectory Error (ATE) and Relative Pose Error (RPE) in both translation and rotation. As shown in the results, our method significantly outperforms the classical VO pipeline. In particular, WAT3R achieves substantially lower ATE and RPE, indicating more accurate and stable camera pose estimation. 

\begin{table}[htbp]
\centering
\caption{\textbf{Comparison of pose estimation accuracy against state-of-the-art SLAM systems.}}
\label{tab:slam_comparison}
\begin{tabular}{l c c c}
\toprule
\textbf{Method} & ATE $\downarrow$ & RPE$_{trans}$  $\downarrow$ & RPE$_{rot}$ $\downarrow$ \\
\midrule
ORB-SLAM3~\cite{ORBSLAM3} & 0.8735 & 0.2703 & 12.4685 \\
\textbf{WAT3R (Ours)} & \cellcolor{bestblue}{0.2321} & \cellcolor{bestblue}{0.0721} & \cellcolor{bestblue}{0.2777} \\
\bottomrule
\end{tabular}
\end{table}

\section{Discussion of Limitations}
\label{sec:limitation}
\begin{figure}[htbp]
  \centering
  \includegraphics[width=0.7\textwidth]{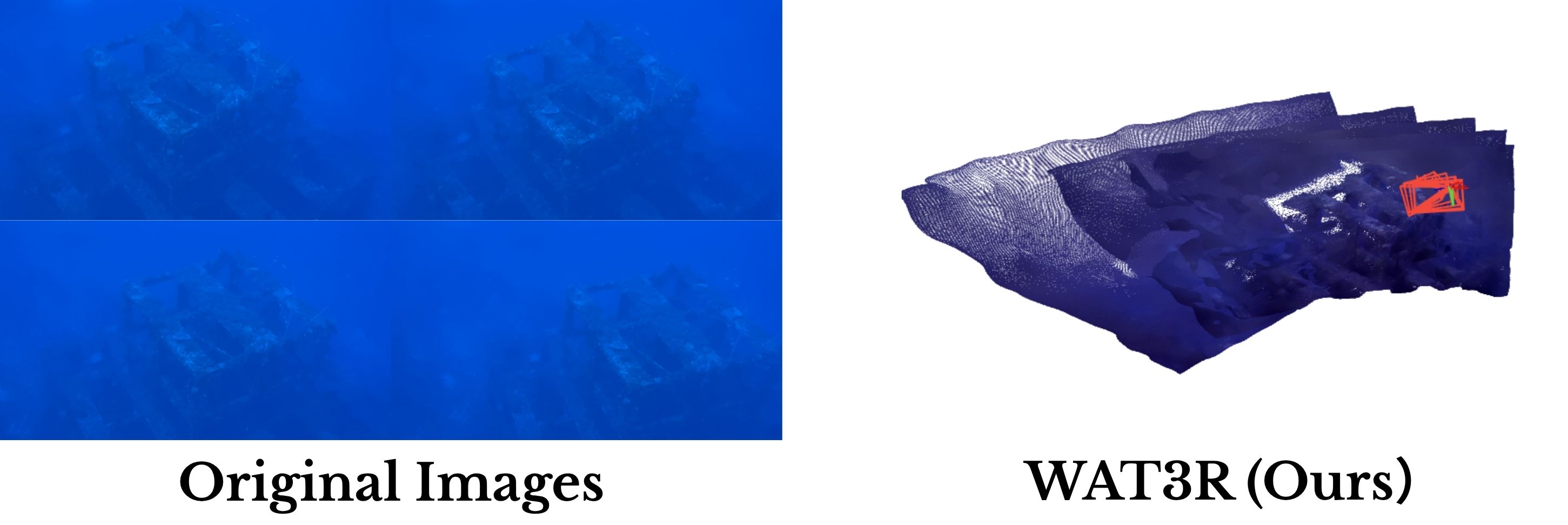}
  \caption{Failure cases of WAT3R under extremely turbid underwater conditions with low visibility.}
  \label{fig:limitation}
\end{figure}

While WAT3R demonstrates reliable 3D reconstruction in challenging underwater conditions, it has one main limitation as shown in ~\Cref{fig:limitation}: extremely turbid environments with very low visibility remain challenging, potentially degrading geometry prediction despite the integrated degradation adaptation signals. In the future, we aim to extend our framework to further improve robustness under highly degraded underwater conditions.
